\documentclass[conference]{IEEEtran}
\IEEEoverridecommandlockouts
\usepackage{url}
\usepackage{amsmath}
\usepackage{amsfonts}
\usepackage{amssymb}
\usepackage{placeins}
\usepackage{subcaption}
\usepackage{booktabs}
\usepackage{graphicx}
\usepackage{float}
\usepackage{csquotes}
\usepackage{dblfloatfix}
\usepackage{array}
\usepackage{tabularx}
\usepackage{hhline}
\usepackage{hyperref}
\usepackage{multirow}
\usepackage{todonotes}
\usepackage[numbers]{natbib}

\newcolumntype{Y}{>{\centering\arraybackslash}X}

\title{Neural Game Engine: Accurate learning of generalizable forward models from pixels.}

\author{
  \IEEEauthorblockN{Chris Bamford}
  \IEEEauthorblockA{\textit{Game AI research group} \\
  \textit{Queen Mary University}\\
  London, UK \\
  \url{c.d.j.bamford@qmul.ac.uk}}

  \and
  \IEEEauthorblockN{Simon M. Lucas}
  \IEEEauthorblockA{\textit{Game AI research group} \\
  \textit{Queen Mary University}\\
  London, UK \\
  \url{simon.lucas@qmul.ac.uk}}

}

\date{\today}

\begin{document}
\maketitle

\begin{abstract}
Access to a fast and easily copied forward model of a game is essential for model-based reinforcement learning and for algorithms such as Monte Carlo tree search, and is also beneficial as a source of unlimited experience data for model-free algorithms.  Learning forward models is an interesting and important challenge in order to address problems where a model is not available.  Building upon previous work on the Neural GPU, this paper introduces the Neural Game Engine, as a way to learn models directly from pixels.  The learned models are able to generalise to different size game levels to the ones they were trained on without loss of accuracy.  Results on 10 deterministic General Video Game AI games demonstrate competitive performance, with many of the games models being learned perfectly both in terms of pixel predictions and reward predictions.  The pre-trained models are available through the OpenAI Gym interface and are available publicly for future research here: \url{https://github.com/Bam4d/Neural-Game-Engine}

\end{abstract}

\section{Introduction}
\label{sec:introduction}

Recent research has focused on forward models of games that can be learned either through heuristic methods or using deep neural network architectures. These learned models can then be used by traditional planning algorithms, or as part of the architecture in reinforcement learning. Using neural networks in planning algorithms can be difficult, as the accuracy of state observations tends to decrease with the number of steps that are simulated. This results in diminishing efficacy of planning algorithms when larger rollout lengths are used.  Recent neural network models also tend to rely on a fixed dimensional observational input to predict the rewards and subsequent states and therefore struggle to generalize to games that may have different sized observational spaces. 

Heuristic rule-based algorithms for learning forward models \cite{dockhorn_2018} offer high performance when they work, but require human input regarding the form the rules will take.  

Recent work on a local approach to learning forward models \cite{dockhorn_2019}
shares some similarities with the Neural Game Engine, in that both methods are able to generalise to levels of a different size than those seen during training.  Compared to \cite{dockhorn_2019}, the Neural Game Engine works directly with pixels rather than tiles, and for many games also does accurate reward prediction.

Grid-based arcade style games, although simple to understand for humans, still present highly challenging environments for artificial intelligence. In this paper a grid-based game refers to a game that is based on a grid of discrete tiles such as walls, floors, boxes and other game-specific items. A single agent has a set of actions it can perform at each time step, such as movement or interaction with other tiles in the grid. The agent is restricted to perform a single action at each time step. Additionally, each environment may have a different grid dimensions, leading to variable observation space sizes. These game environments can be represented by a fully observervable markov decision process with states $s$ as the pixels of the environment, actions $a$ of the agent and the rewards $r$ given by the game score.

This paper proposes a novel architecture, the Neural Game Engine, based on a modified {\em Neural GPU}\cite{kaiser_2015} \cite{freivalds_2017}.

The Neural Game Engine (NGE) can learn grid-based arcade style games of any dimensions with very high accuracy over arbritary numbers of game ticks. Additionally the architecture can scale to grid games of any number of tiles without loss of accuracy. The NGE engine is trained on several deterministic games \footnote{Pre-traied models are available through the OpenAI Gym interface and are available publicly for future research here: \url{https://github.com/Bam4d/Neural-Game-Engine}} from the GVGAI environment \cite{perezliebana_2018}, an updated version of pyVGDL \cite{schaul_2013} which provides many grid-based games under the openAI gym wrapper \cite{open_ai_gym}.

The paper is structured as follows: Firstly, section \ref{sec:background} covers recent similar research that covers the generation of forward models and how they have been used in various research such as statistical forward planning algorithms or reinforcement learning.

Section \ref{sec:NGPU} describes the {\em Neural GPU} in detail and the modifications it requires to learn high accuracy game models. Evaluation methodology and training details are given in sections \ref{sec:evaluation_methodology} and \ref{sec:training}. Finally, results of various architecture and training experiments are given in section \ref{sec:results}

\section{Background} \label{sec:background}

\subsection{Learning Forward Models} \label{sec:learning-forward-models}

Humans have the ability to be able to model the outcome of their actions. This is achieved through having an internal model of the environment in which different actions can be tested out and appropriately chosen. This inherent ability allows humans to perform tasks such as planning or seeking intrinsic rewards \cite{schmidhuber_2010}.
For artificially intelligent agents, having access to, or learning the model of its environment through experience is arguably an unavoidable step towards being able to achieve human-level intelligence.

Deep neural networks have been used successfully to estimate forward models for various use cases. For example {\em curiosity driven exploration} \cite{pathak_2017} make use of having forward model that can be used to measure uncertainly about particular states. This measure of uncertainly is used as intrinstic motivation to drive the agent to explore actions that take the agent to states that have not been seen before.

Similarly predictions about how much information the agent has access to in certain states have been used to try and maximise the {\em Empowerment} of the agent. If an agent is more empowered, it typically has greater access to states that will end in high rewards \cite{houthooft_2016}.

In many cases, it is very difficult to learn a model of the environment which can perfectly reproduce the environment dynamics and observation frames over many time steps, especially if the model has stochastic elements such as enemies that move in unpredictable ways.

In \cite{ha_2018}, a combination of auto-encoders and recurrent neural networks are used to predict the next frames of several OpenAI gym games. Auto-encoders were used to encode the data of a single frame into a latent state $z_t$, this latent state is then used in combination with an LSTM, which stores information about previous states, to output a probability density function $P(z_{t+1} | a_t, z_t, h_t)$ where $a_t$ is the action applied at time $t$ and $h_t$ is the hidden state produced by the previous LSTM cell. This distribution can then be used to produce the next frame.

More recently generative models have been used to predict frames of environments by sampling from a distribution $p(s'|s,a)$ where $s'$ is the state being predicted. Generative models allow the capture of stochastic and deterministic dynamics of game states, and can even predict the actions of NPC based characters.

Generative state space models \cite{buesing_2018} \cite{gregor_2019} \cite{hafner_2018} \cite{hafner_2019} encode state information into a typically 3-dimensional tensor instead of a latent vector allowing a richer representation of the underlying environment states, these models are commonly combined with recurrent neural network techniques such as LSTMs and GRUs and are generally more accurate than latent vector encoding of states. State space representations are also used in \cite{gregor_2018} in order to predict future states without performing step-by step rollouts.

\subsection{Local Forward Modelling}

A recent successful method of learning forward models of grid-based games is to use {\em Local Modelling}. Local modelling takes advantage of the fact that in many games, the mechanics can be applied to small areas of the game environment independently of others. The most simplistic example of a local model is that of a 2D cellular automata. It has been shown that using local modelling, forward models of basic cellular automata based games can be learned by focussing on the rules which modify the state of single tiles based on the surrounding tiles \cite{lucas_2019} \cite{dockhorn_2019}. In \cite{weber_2017} the model used for learning the forward model of the imagined Sokoban game is the equivalent of a local feed-forward cellular automata model. This technique is also used in \cite{buesing_2018} \cite{hafner_2019} for state-space transitions and during encoding and decoding of pixels to the latent state-space.

Local modelling is also used in \cite{guez_2019} as part of a model-free RL agent. Because this architecture works well with tasks that usually require agents to plan, it is argued that although this architecture is not explicitly trained to reproduce the underlying environment model, it is learning to plan implicitly.

\subsection{Neural GPU} \label{sec:NGPU}

The Neural GPU (NGPU) architecture introduced in \cite{kaiser_2015} and improved upon in \cite{freivalds_2017} can learn several unbounded binary operations. For example, multiplication, addition, reversal and duplication. This is achieved by effectively learning 1D cellular automata rules which are then applied over a number of steps until the result is achieved. The number of steps $I$ is typically proportional to the size of the binary digits being processed. The Neural GPU applies the cellular automata rules to an embedded representation of the binary digits using a convolutional gated recurrent unit (CGRU) with hard non-linearities. The CGRU itself is described by the following set of update rules:

\begin{equation} \label{eq:CGRU}
	\begin{aligned}
		 & u_i = \hat{\sigma}(U' * s_{i-1} + B')         \\
		 & r_i = \hat{\sigma}(U'' * s_{i-1} + B'')       \\
		 & c_i = \hat{tanh}(U * r_i \odot s_{i-1} + B)   \\
		 & s_{i} = u_i \odot s_{i-1} + (1-u_i) \odot c_i \\
	\end{aligned}
\end{equation}

In the above equations $U$, $U''$, $U'''$ are convolutional kernel banks and $B$, $B'$, $B''$ are learnable biases. The $*$ operator is used to describe a convolution operation of the left parameter over the right. For example $U' * s$ denotes the kernels in $U'$ convolved over the values in $s$. $\hat{tanh}$ and $\hat{\sigma}$ represent the hard non-linearity versions of the {\em tanh} and {\em sigmoid} functions respectively and $\odot$ represents the Hadamard (or element-wise) product between two tensors. Details of the hard non-linearities are given in the original paper \cite{freivalds_2017}.
When dealing with binary operations, the Neural GPU takes an input of arbitrary length, containing the binary encoded digits and the operation to perform. The binary digits and operation symbol are embedded into the initial state $s_0$ this state is then iterated through the CGRU for $n$ steps and final state $s_n$ is read out using a softmax layer which predicts the binary result.

As the Neural GPU can be seen as a recurrent application of learnable cellular automata rules, this leaves it well suited to being able to learn the local rules of grid-world based games. This architecture is comparable to other state-space architectures that use size-preserving layers \cite{weber_2017}, \cite{buesing_2018} \cite{guez_2019}, with the exception that parameters are shared between layers, no latent state information is shared between frames and different gating mechanisms are explored.

\section{Neural Game Engine}
\label{sec:NGE}

\begin{figure*}
	\centering
	\includegraphics[width=\textwidth]{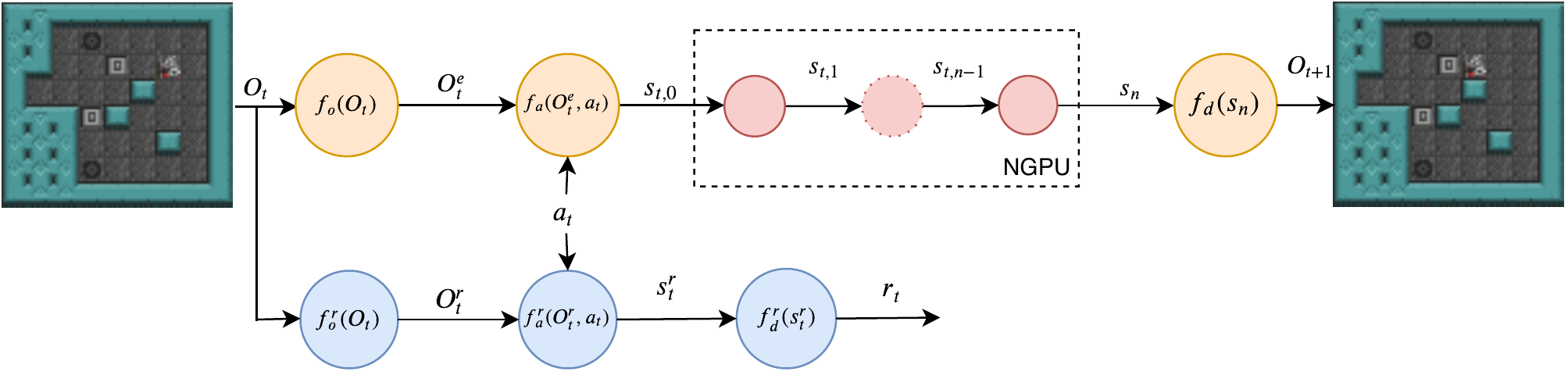}
	\caption{Architecture of the Neural Game Engine}
	\label{fig:nge_architecture}
\end{figure*}

The Neural Game Engine is a neural network architecture with a modified Neural GPU at its core. The main modifications to the Neural GPU are outlined in this section. In the Neural Game Engine the state $s$ takes a two dimensional shape $(W_s, H_s, C_s)$ where the width $W_s$ and height $H_s$ reflect the width and height in tiles of the game being trained and $C_s$ is the number of channels. Each vector stored at $(w_s, h_s)$ represents a single tile in the grid environment. The convolutional kernel banks $U$, $U'$ and $U''$ are also modified to be two dimensional with a shape of $(3,3)$. The stride and zero-padding are kept the same as the original paper at $1$. As there are no diagonal movements allowed in any GVGAI environment, the kernels are also masked to ignore the non-adjacent cells.
Similarly to the NGPU, an iteration of the CGRU unit with input $s_i$ produces a new state $s_{i+1}$. The number of iterations of CGRU cell per frame of the game state is tuned as a hyperparameter $n$.

The width and height of the games in the GVGAI environment can be any positive integer value. Due to the fact that changing the values of $W_s$ and $H_s$ does not result in any change of the number of parameters in the  underlying Neural GPU, this means that the Neural Game Engine can generalize to any $W_s$ and $H_s$. This unbounded computation of game state is discussed further in section \ref{sec:unbounded}.

In many reinforcement learning techniques, the rewards that the game provides to the player are augmented in order to aid exploration, modify the agent's goals, or provide auxillary losses to reduce training time \cite{shelhamer_2016}, \cite{andrychowicz_2017}. In some cases the original rewards supplied by the environments are modified from their original values with a technique known as reward shaping \cite{grzes_2017}.

Reward prediction in the Neural Game Engine aims to reproduce the original game rewards as accurately as possible, but taking into account that reward prediction should be a seperate process entirely from the game mechanics.

At every time step the Neural GPU is applied to an encoded observation image $O_t$ and iterated $n$ times to  and then decoded to give the next observation state $O_{t+1}$ and reward $r_t$.

The architecture for a single time-step calculation is shown in figure \ref{fig:nge_architecture}.

\subsection{Observation Encoder - \boldmath$f_o(O_t)$}

In the GVGAI environment, tiles in the trained games are set to have the same width and height dimensions $D$. This consistency allows the tiles to be embedded into a tensor $O^e_t$ with the same dimensions of the NGPU initial state $s_0$. This tile embedding is achieved by using a convolutional neural network with kernel width, kernel height and stride set to $D$, input channels set to $3$ to reflect the RGB components of the image and finally output channels set to $C_s$, the number of channels in the NGPU state.

\subsection{Observation Decoder - \boldmath$f_d(s_{n})$}

To render the game pixels, a mapping from the underlying embedded tile representations to the pixel representations of the tile is learned. This mapping takes the form of a convolutional transpose with kernel size $D$ and stride $D$. The number of input channels is set to 3 to reproduce the RGB components. This mapping recovers a tensor of shape $(D.W_s, D.H_s, 3)$ which can be rendered.

\subsection{Action Conditioning - \boldmath$f_a(O^e_t, a_t)$}

As the action needs to be considered as part of the local rule calculations in the NGPU, information about the actions must be available in the $s_0$ state, along with the observations. To achieve this, the action $a_t$ is one-hot encoded and then embedded with a linear layer of output size $C_s$. This is then added to each cell of the initial state $s_0$. In practice this can be achieved by tiling the one-hot representation of the action into a tensor of size $(W_s, H_s, A_s)$, where $W_s$ and $H_s$ are the width and height of the NGPU state, and $A_s$ is the cardinality of the set of actions for the game. This state can then by passed to a 1x1 convolutional neural network with $C_s$ output channels. The resulting tensor can then be added to the $O^e_t$, which results in the initial $s_0$ state of the Neural GPU.

\subsection{Reward Observation Encoder - \boldmath$f^r_o(O_t)$}

The reward observation encoder consists of a tile embedding layer similar to the observation encoder encoding each tile into a vector with $C_r$ channels, giving an embedded observation state $O^r_t$ of size $(W_s, H_s, C_r)$.

\subsection{Reward Action Conditioning - \boldmath$f^r_a(O^r_t, a_t)$}

A seperate action conditioning network encodes the action at each step $a_t$ to a one hot vector which is the embedded into a linear layer of size $C_r$ and then added to each of the embedded tile vectors giving the reward state $s^r_t$. This process is identical to the NGPU action conditioning, the only difference is the number of channels may be different depending on hyperparameter choices.

\subsection{Reward Decoder - \boldmath$f^r_d(s^r_t)$}

In order to decode the rewards from the reward state $s^r_t$, a convolutional network network with kernel size of 3 and padding 1 is used followed by two convolutional layers with kernel size of 1, 0 padding and number of channels decreasing in each layer. A final convolutional layer with kernel size 3 is used to decrease the  number of channels to 16 and an arbitrary height and width. Global max pooling is applied across the remaining arbitrary height and width dimensions leaving 16 outputs. These 16 ouputs are then trained with categorical cross entropy loss to predict an 8 bit binary number corresponding to the reward. Predicting binary rewards in this way instead of predicting linear values means that reward prediction is reduced to a classification problem. This has the effect that variance in underlying parameters can remain low. Negative reward values given in the original environment are currently ignored as the predicted binary number is unsigned. To support negative rewards, a sign bit or two's complement encoding could be used. To predict fractional rewards, float or double encoding could be used.

\section{Neural GPU enhancements}

\subsection{2D Diagonal Gating}

Diagonal gating, introduced in \cite{freivalds_2017} is a technique used in the NGPU architecture to allow state cell values to be passed directly to neighbouring states cells. In the context of a grid-world game, it follows that information such as tile type, could be transferred in this manner. The state of the original NGPU is a one-dimensional vector and thus its diagonal gating mechanism allows it to copy state information from the left and right cells. The state of the underlying NGPU in the Neural Game Engine is two dimensional, which means that the diagonal gating mechanism can copy from above and below, as well as left and right. To achieve this, the state is now split into 5 parts $s_{i} = (s^1_{i},s^2_{i},s^3_{i},s^4_{i},s^5_{i})$ and a 2D convolution operator with fixed kernels as shown in equation \ref{eq:2d-diag-gating} is used.

\begin{equation} \label{eq:2d-diag-gating}
	\begin{aligned}
		 & s_i = u_i \odot \tilde{s_i} + (1-u_i) \odot c_t                                                 \\
		 & \tilde{s_i} = (\tilde{s^1_{i}},\tilde{s^2_{i}},\tilde{s^3_{i}},\tilde{s^4_{i}},\tilde{s^5_{i}}) \\
		 & \tilde{s^1_{i}} = s^1_{i-1} * [
				[0, 1, 0],
				[0, 0, 0],
				[0, 0, 0]
			]                                                                                                  \\
		 & \tilde{s^2_{i}} = s^2_{i-1} * [
				[0, 0, 0],
				[0, 0, 1],
				[0, 0, 0]
			]                                                                                                  \\
		 & \tilde{s^3_{i}} = s^3_{i-1} * [
				[0, 0, 0],
				[0, 0, 0],
				[0, 1, 0]
			]                                                                                                  \\
		 & \tilde{s^4_{i}} = s^4_{i-1} * [
				[0, 0, 0],
				[1, 0, 0],
				[0, 0, 0]
			]                                                                                                  \\
		 & \tilde{s^5_{i}} = s^5_{i-1} * [
				[0, 0, 0],
				[0, 1, 0],
				[0, 0, 0]
			]                                                                                                  \\
	\end{aligned}
\end{equation}

\subsection{Selective Gating} \label{sec:selective-gating}

One of the issues with diagonal gating is that the copying of the state information is uni-directional for the state values in each cell $(w_s, h_s, c_s)$. To illustrate this issue consider the values in any sub-state $s^x_{i}$. The values in each sub-state are only shifted in a single direction. This means that sub-states that are shifted in one direction are not the same states that can be shifted in the other directions. This uni-directional flow does not allow consitent copying of state information across all directions. Intuitively this means that if a tile moves upwards, the state information it can bring to the cell above cannot be moved to the left, right or even back to the cell that it started in.

To alleviate this issue, a {\em selective gating} mechanism is proposed which allows the gating mechanism to copy values in any direction for any value in any cell $(w_s, h_s, c_s)$.

The selection mechanism works by learning a classifier that, given the state tensor $s_i$ outputs a selection tensor $\hat{S}$ of dimensions $(W_s, H_s, C_s, 5)$ where the selection of the gating directions (up, down, left, right, center) are one-hot encoded into the last dimension. The selection tensor is created by applying a convolution operation to the state $s_i$ with kernel size of 3x3, stride of 1, padding of 1 and $5C_s$ output channels. The $5C_s$ channels are then reshaped into a tensor of size $(5,C_s)$ and a softmax applied across the first dimension to give a {\em selection} for each of the $C_s$ values.
The selection tensor $\hat{S}$ is then multiplied by a tensor $\hat{K}$ of shape $(5,C_s,1,W_s,H_s)$ containing 5 directionally shifted versions of the original state. This gives the new state $\tilde{s_i}$.

\begin{equation} \label{eq:selective-gating}
	\begin{aligned}
		 & s_i = u_i \odot \tilde{s_i} + (1-u_i) \odot c_t               \\
		 & \tilde{s_i} = \hat{S} \hat{K}                                 \\
		 & \hat{K} = [M_{u}(s_i),M_{d}(s_i), M_{l}(s_i),M_{r}(s_i), s_i] \\
	\end{aligned}
\end{equation}

The shifting operation can be achieved by the convolution of a fixed kernel that copies states from adjacent cells. Zero padding of 1 is applied so the state retains its original shape. For example:

\begin{equation} \label{eq:selective_M}
	\begin{aligned}
		 & M_{u}(s_i) = s_i \odot [[0, 1, 0],[0, 0, 0],[0, 0, 0]] \\
		 & M_{d}(s_i) = s_i \odot [[0, 0, 0],[0, 0, 0],[0, 1, 0]] \\
		 & M_{l}(s_i) = s_i \odot [[0, 0, 0],[1, 0, 0],[0, 0, 0]] \\
		 & M_{r}(s_i) = s_i \odot [[0, 0, 0],[0, 0, 1],[0, 0, 0]] \\
	\end{aligned}
\end{equation}

\subsection{Evaluation Methodology} \label{sec:evaluation_methodology}

The aim of the experiments is to try reach pixel-perfect reproduction of original GVGAI environment games over abitrarily long time frames for levels with any dimensions. In order to achieve this, the network must learn the game mechanics on a symbolic level and then be able to apply these to larger game states.

The results presented in this paper are performed on the game {\em Sokoban} as it is a good example of a GVGAI game with local rules.

In order to measure the accuracy of the reproduction of the game, two related measures are used. Firstly the mean-squared error $E_{mse}$ of the raw pixel outputs at each step and secondly, a {\em closest tile f1} $F_t$ measure. The {\em closest tile measure} is created by firstly taking a tile map of the original observation $T_m$ which has dimensions $(W_s, H_s)$ where each element in the map corresponds to an index of the set of possible tiles $\mathcal{T}$. A second tile map $\hat{T}_m$ is then created by finding the closest matching tile in the set of tiles $\mathcal{T}$ for each $D$x$D$ tile in the predicted observation. The {\em closest tile f1} measure is calculated from the mean of the f1 scores for each of the tiles in $T_m$ and $\hat{T}_m$. The f1 scores are generated by measuring the precision and recall of the tile predictions.

Alongside learning the pixel-accuracy, the rewards given by the environment are learned. Reward error is measured by converting the real reward values to a binary representation and then calculating the cross entropy loss. Reward accuracy is measured using precision, recall and f1 $F_r$ score of the binary classifications.

\subsection{Training} \label{sec:training}

In order to obtain accurate rollouts over long time periods, for any size network, the training data is generated in a way that does not bias towards game sizes, numbers of tiles (such as walls, boxes and holes in sokoban), or particular RL or planning policies.

Level generation for GVGAI games has been explored in \cite{khalifa_2016}, \cite{drageset_2019} and \cite{justesen_2018}. However these generators are aimed at either producing levels that help RL agents to learn or are pleasing to human players.

To generate levels for learning the environment dynamics, the probability for an agent to interact with different types of tiles must be evenly distributed. To achieve this, levels are randomly generated with height $H$ and width $W$ between certain values $H_{min}$, $H_{max}$, $W_{min}$, $W_{max}$. GVGAI environments typically contain 5 pre-built levels. These pre-built levels are used to generate the probabilities of each tile being placed in the environment. Tiles are positioned with these calculated probabilities with the caveat that wall tiles are always placed on the edges of the game state if this is consistent with the 5 pre-built levels. Additionally, tiles that only appear once in each level are placed only once in generated levels.

A random agent is used to generate experience data in the environment. To improve the distribution of training data, each step is augmented by creating 8-way tile-symmetrical observation and actions. Each step of learning uses mini-batch gradient descent, where the batch contained the symmetrical experiences. Batch sizes are fixed at 32 state transitions, giving a total of 256 frame transitions per batch. Similarly to the Neural GPU, saturation cost is calculated for the hard-non linearites in the CGRU units, however these saturation costs are averaged across the batch instead of summed, which increased the stability of training and improved training results overall. The saturation limit in all experiments is set at 0.99 and weighted at 0.001. Saturation cost is not clamped at any value with respect to the overall loss as it is in the original paper.

As the observation predictions at each time step become the inputs for the next prediction, errors can build up over time and cause the rollout accuracy to decrease rapidly. During experiments, the same Prediction Dependent Training (PDT) technique introduced in \cite{chiappa_2017} coupled with a curriculum schedule was employed which increased accuracy and training stability. Observation noise is also added to training data, this was integral to acheiving high accuracy.

In order to evaluate the training progress of the environment, rollouts are performed every 200 epochs using real game levels from the GVGAI environment. 3 repeats of rollouts of length 100 are performed, and $F_t$ and $E_{mse}$ are calculated.
\footnote{All training and testing is performed on a single Ubuntu 18.04 machine with an NVIDIA 2080ti GPU, Intel® Core™ i7-6800K CPU and openBLAS (0.2.20) libraries installed.}

\section{Experiments and Results} \label{sec:results}

\subsection{Comparison of gating mechanisms}

\begin{figure}
	\includegraphics[width=\linewidth]{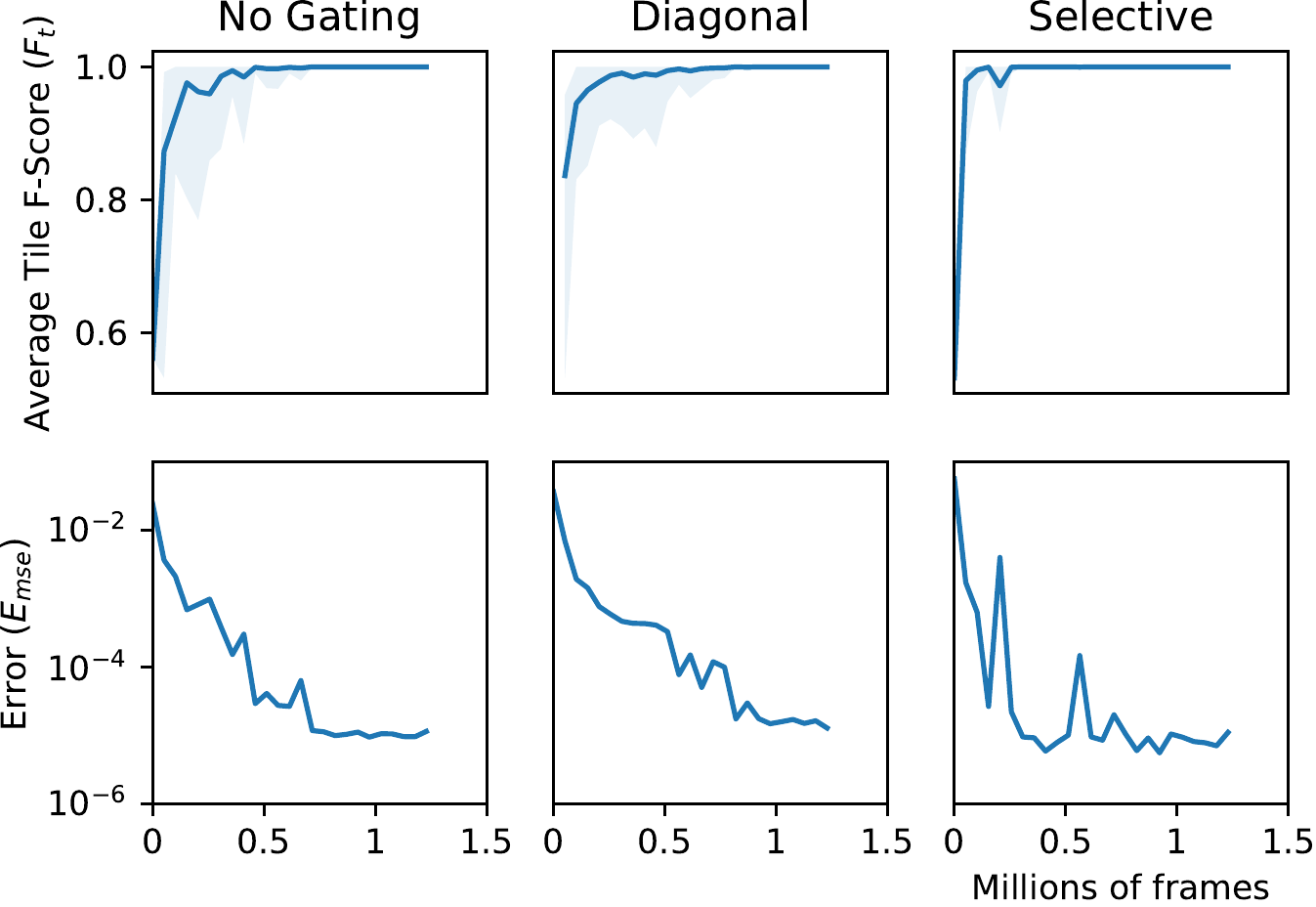}
	\caption{F1 score ($F_t$) and mean squared error ($E_{mse}$) when training the three gating mechanisms over 1.28M frames. Selective gating trains fastest and is the most accurate.}
	\label{fig:training_compare_gating}
\end{figure}

In figure \ref{fig:training_compare_gating}, the NGE architecture using the different NGPU gating mechanisms described in \ref{sec:NGE} is shown. Even with no diagonal or selective gating, the NGE can very learn very accurate models of game environments. In the experiments, {\em Selective} gating had a small advantage in stability over long time horizons, this is also reflected in table \ref{tab:unbounded}.

\subsection{Comparison with other methods}

The best Neural Game Engine (NGE) model is compared against several common networks from recent literature with the game Sokoban. Rewards prediction is not analysed as it is a seperate network. The The network architectures that are compared are the following:

\paragraph{\textbf{FeedForward (FF)}} This model replaces the NGPU module with two feed-forward convolutional layers with kernel size of 3, stride 1 and padding 1. This is the equivalent of the  {\em basic block} used in \cite{weber_2017} when training Sokoban. The model compared does not use pool-and-inject layers as Sokoban has no long-distance dependencies that require global state changes. This model is commonly used as the determinsitic component of generative state-space architectures and is well suited to deterministic grid environments.

\paragraph{\textbf{Recurrent Environment Simulators (RES)}} The state of the game is encoded into a latent state using an auto-encoder. This latent state then forms the input to an LSTM unit which can store past state information in its hidden state. This model is equivalent to the Recurrent Environment Simulator (RES) \cite{chiappa_2017} and models that use an auto-encoder to create a latent state.

\paragraph{\textbf{Stochastic State Space (sSS)}} The most complex model which, like NGE heavily uses cellular automata-like layers which encode pixel information into a compressed grid. The model differs from NGE in that it works with continuous and stochastic environments, and therefore uses sampling in order to produce the output observations.

\begin{figure}
	\centering
	\includegraphics[width=\linewidth]{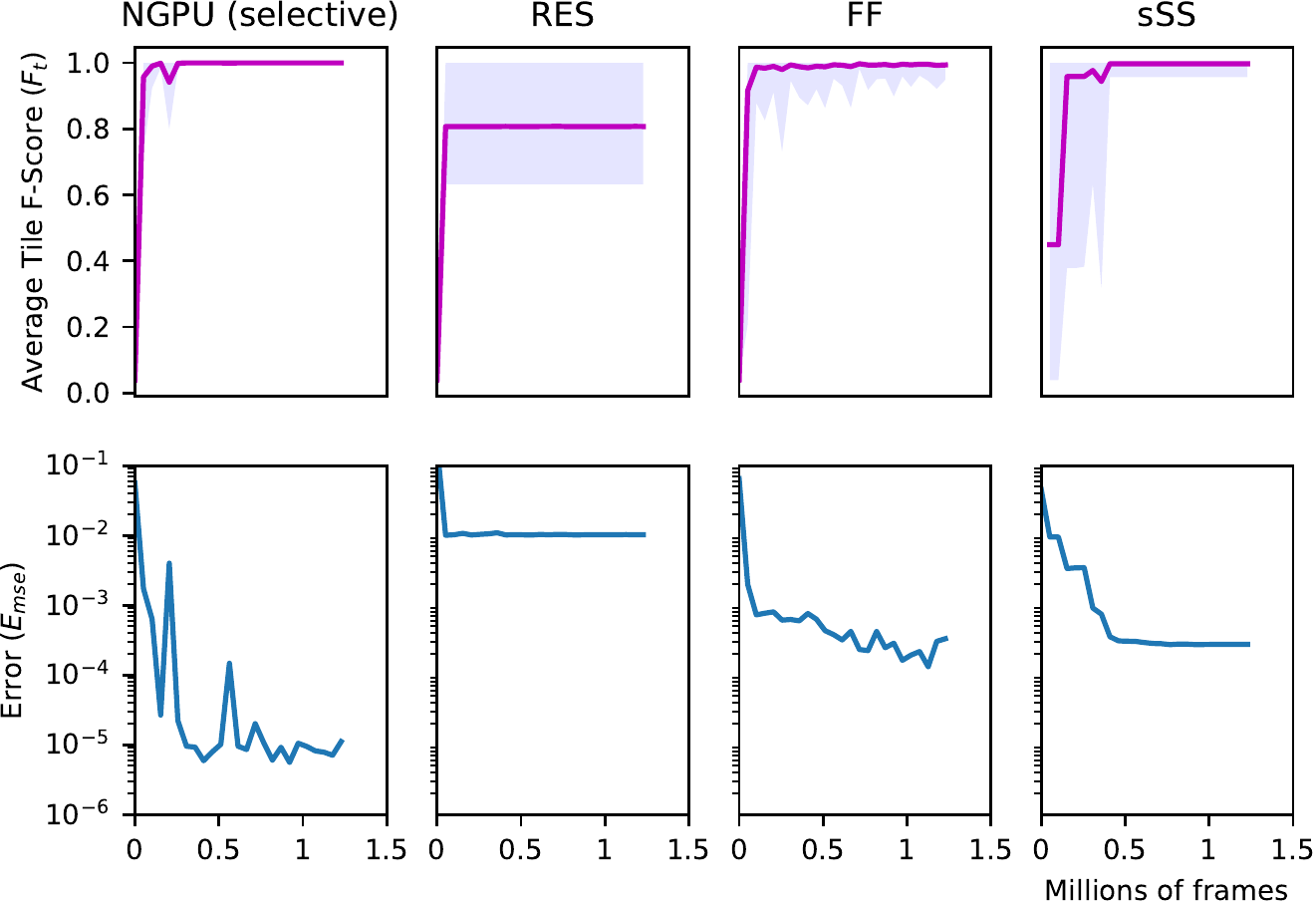}
	\caption{F1 score ($F1_t$) and mean squared error ($E_{mse}$) when each forward model method is trained over 1.28M frames. NGPU acheives the lowest $E_{mse}$ and higest average $F_t$ score.}
	\label{fig:training_compare_methods}
\end{figure}

Figure \ref{fig:training_compare_methods} shows the comparison of these 4 methods with the same input data and number of epochs. The training in this experiment is limited to random grids of fixed size (10x10). This is due to the fact that RES and sSS models contain architectural components that cannot generalize to different size grids. Each method trains to high accuracy very quickly, followed by a plateau in decreasing error, leading to a maximum accuracy. In the case of Sokoban, FF, sSS and NGPU methods have a slight advantage as Sokoban is naturally suited to local modelling. However the sSS model is disadvantaged by the fact that it contains stochastic components that are trying to model completely deterministic state transitions.

\subsection{Ablation Testing}

An important feature of many games is that many interactions between adjacent cells can be dependent on other surrounding cells. For example in Sokoban, when pushing a movable block against a wall, a 3x3 grid around the location of the agent will not take into account the wall when calculating the next state of the cell currently occupied by the agent. However, the NGPU accounts for non-local interactions when it iterates during a single time step. This effectively lets cells share information during the processing of a single state. With $n = 2$, the NGPU can share information from more adjacent cells, encompassing the wall that the block cannot be moved past. Other models such as those used in \cite{buesing_2018} \cite{weber_2017} use similar techniques, but use fixed networks with different convolutional network sizes and apply residual layers. Using a NGPU with multiple iterations removes the requirement for multiple layers of convolutions and residual connections, making the network much simpler and smaller.

In \cite{freivalds_2017}, diagonal gating is used to share state information between adjacent cells. As described in section \ref{eq:selective-gating} this only allows single-direction information flow, which reflects in the higher error rate of NGE models using diagonal gating.

\begin{figure}
	\includegraphics[width=\linewidth]{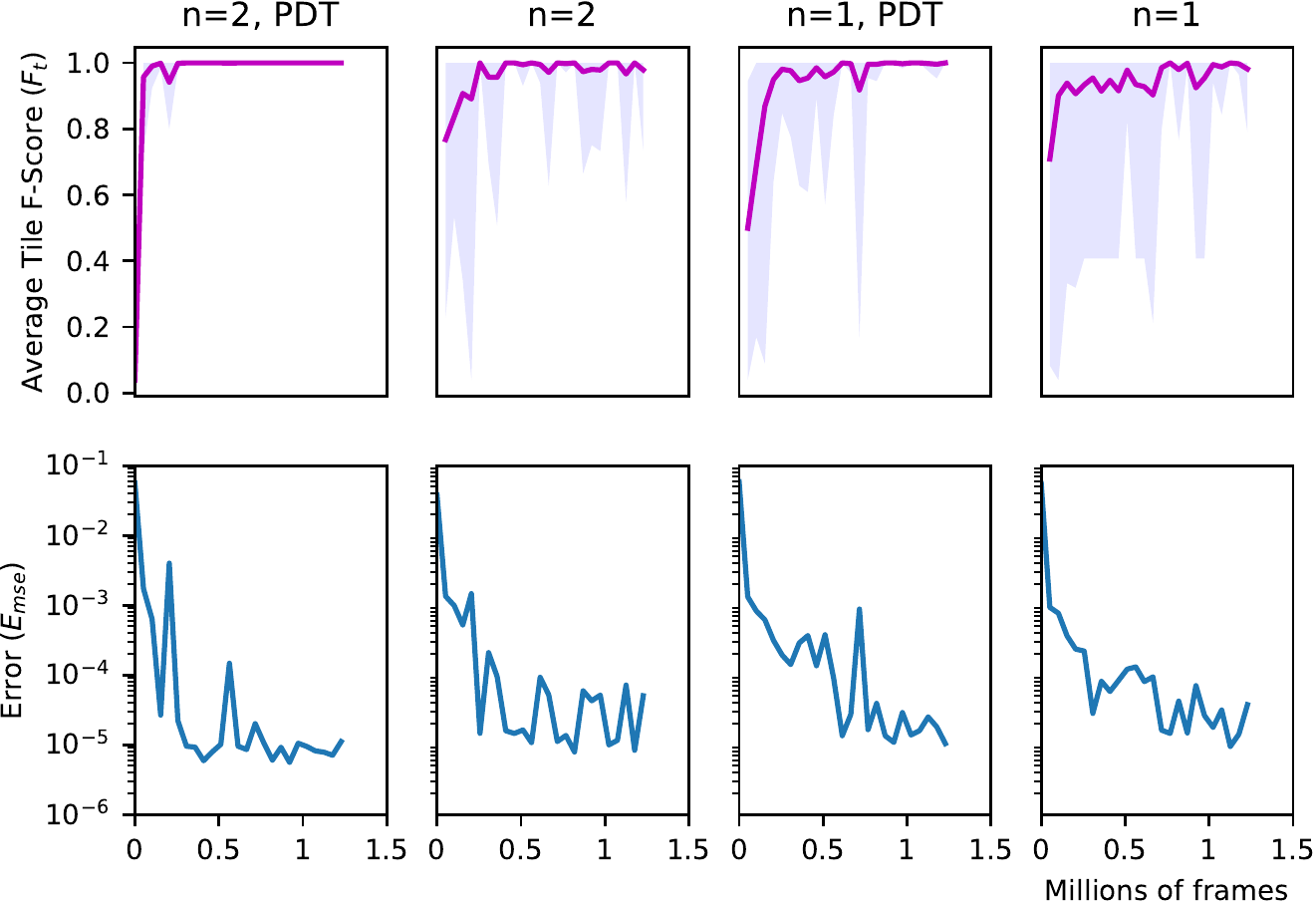}
	\caption{Learining a very accurate model of the Sokoban environment is dependent on having multiple iterations of the NGPU units and also training over multiple states. When the network is restricted to a single iteration $n = 1$ or is trained without PDT, the accuracy suffers. The results shown here are the rollout accuracy measurements against the 5 original hand-built GVGAI environments, taken every 200 epochs during training.}
	\label{fig:training_compare_ablation}
\end{figure}

To test that the iteration of NGPU is vital for information flow in local interactions, two experiments are performed under all the same conditions of the high performing models. One with the modification that only a single NGPU step and no PDT is configured during training. The other with a single NGPU step, but using PDT described in section \ref{sec:training}. The second experiment aimed to rule out that local information could be transported through pixels. The results of this are shown in figure \ref{fig:training_compare_ablation}.

In both the 2-step and 2-step+PDT experiments, the accuracy achieved is very high, but with the single step options, the accuracy achieved plateaus at a much lower value and the prediction error remains high. This result shows that multiple steps of the NGPU are vital to achieving high accuracy. It's also important to note that the 2-layer FF model in figure \ref{fig:training_compare_methods} also could not achieve this high accuracy.

\subsection{Unbounded Computation} \label{sec:unbounded}

To test the generalization ability of the trained NGE, the models trained in section \ref{sec:training} are used to play several levels with much larger dimensions than those during training. These larger models are then compared against the the original GVGAI environment with an identical starting state and action list. The two methods ($E_{mse}$ and $E_{t}$) of measuring the accuracy of these models are used as described in section \ref{sec:training}.
For each model, the two measure are calculated for each step up to 500 and an average of the measures are taken over 10 repeats. These results are shown in table \ref{tab:unbounded}

\begin{table}
	\begin{center}
		\begin{tabular}{||c c c||}
			\hline
			Grid Size & $max(E_{mse})$ & $F_{1}$      \\ [0.5ex]
			\hline\hline
			30x30     & 7.5e-6         & \textbf{1.0} \\
			\hline
			50x50     & 8.3e-6         & \textbf{1.0} \\
			\hline
			70x70     & 7.9e-6         & \textbf{1.0} \\
			\hline
			100x100   & 7.9e-6         & \textbf{1.0} \\
			\hline
		\end{tabular}
	\end{center}
	\caption{The maximum mean squared error and closest tile error for 500 steps averaged over 10 repeats. NGPU with {\em Selective} gating obtains high tile accuracy in all of the generalization tests.}
	\label{tab:unbounded}
\end{table}

\subsection{Results on GVGAI games}

The results of training Neural Game Engine on several GVGAI games is shown in table \ref{tab:gvgai_results}. The rollouts follow the same setup described in section \ref{sec:evaluation_methodology} Games that result in $F_{t}$ scores of \textbf{1.0} show that the underlying game rules are learned accurately and the NGE does not make any mistakes when tested. Reward F1 scores $F_{r}$ can be interpreted in the same way. Most of the tested games acheive high accuracy, however there are some game mechanics that cannot be supported by the NGE without modifications. As an example, {\em clusters} completely fails to learn the reward function.  Rewards are fairly common in the game and the forward model itself learns very accurately, so the reason for this is unclear. The game {\em aliens} is included as an example of a game that has stochastic (the enemies randomly shoot at the player) and partially observable (the enemies spawn from a location that has no visible markers) components. The reward function $min(F_{r})$ of aliens is partialled learned by NGE, however the $min(F_t)$ score is 0.73 meaning that just over a quater of the tiles are predicted incorrectly.

\begin{table}
	\begin{center}
		\begin{tabular}{||c c c c||}
			\hline
			Game        & $max(E_{mse})$ & $min(F_{t})$ & $min(F_{r})$ \\[0.5ex]
			\hline\hline
			sokoban     & 7.5e-6         & \textbf{1.0} & \textbf{1.0} \\
			\hline
			cookmepasta & 9e-4           & 0.98         & 0.83         \\
			\hline
			bait        & 5.2e-4         & 0.97         & 0.99         \\
			\hline
			brainman    & 3.6e-4         & 0.97         & \textbf{1.0} \\
			\hline
			labyrinth   & 1.6e-5         & 0.97         & \textbf{1.0} \\
			\hline
			realsokoban & 1.8e-3         & 0.86         & \textbf{1.0} \\
			\hline
			painter     & 4.6e-6         & \textbf{1.0} & \textbf{1.0} \\
			\hline
			clusters    & 1.3e-5         & \textbf{1.0} & 0.0          \\
			\hline
			zenpuzzle   & 8.2e-6         & \textbf{1.0} & \textbf{1.0} \\
			\hline
			aliens      & 5.1e-3         & 0.73         & 0.85          \\
			\hline
		\end{tabular}
	\end{center}
	\caption{The maximum mean squared error $max(E_{mse})$, minimum closest tile f1 $min(F_{1})$ and minimum reward f1 $min(F_{r})$ for 100 steps over 3 repeats.}
	\label{tab:gvgai_results}
\end{table}

\section{Discussion}

There are several interesting applications for games trained with the NGE architecture, for example the fact that games can be learned with very high accuracy over long time horizons, these can be used in planning algorithms. Additionally, because these games also run entirely on the GPU, the sample rate and parallelization ability mean that they can be used as efficient environments for reinforcement learning experimentation.

There are two main limitations that the NGE architecture suffers from: its lack of ability to model stochastic game elements and global state-changes. Further experimentation and research is required to achieve these goals. One approach could be to use NGPU modules in place of the deterministic size preserving layers in sSS models.

One large area for improvement for the Neural Game Engine is that the statical method of level generation and random agent movement does not produce enough examples for some local patterns. In many cases, tweaking the random level generation parameters is enough to give the NGPU a distribution which greatly improves the accuracy of training. Improving the data distribution of local states to train the NGPU is an area which could greatly be improved. Using curiosity driven agents, or planning agents may provide much better data distributions for learning rewards, but may avoid areas of low rewards and therefore not learn the full game dynamics.

Another area of improvement would be that the Neural Game Engine only predicts a single time step in the future, therefore events that do not specifically change the observational state are completely lost. For example, in some games the agent picks up a key and then the agent tile changes to show the agent holding a key. Once the agent has a key, the agent can open a door. NGE learns these dynamics well and learns that if the agent lands on a tile with a key, it changes to an agent with a key and can then interact with a door. However if the fact that the agent is holding a key does not change the agent tile, NGE has no knowlege of this at the next step and therefore the information is lost. This could be fixed by following the latent state space model training techniques used in \cite{buesing_2018}, \cite{amos_2018} and \cite{azar_2019} where future observations are predicted several steps in the future without decoding the visual information between steps.

\section{Conclusion}

In this paper, the Neural Game Engine architecture is proposed as a method of learning very accurate forward models for grid-world games. The Neural Game Engine architecture, which is built upon the Neural GPU, learns a set of underlying local rules that can be applied over several iterations rather than stacking layers with different parameters. Improvements to the Neural GPU architecture such as selective gating are introduced which enable it to be applied to predicting the forward dynamics of games. This paper shows that this method has many advantages: fast learning time; very high accuracy over long time-horizons and fast and easily parallelised execution. The Neural Game Engine shows higher accuracy at predicting the state transitions in the game Sokoban when compared to similar state space models that are used in several model-based reinforcement learning applications. Additionally the Neural Game Engine is shown to generalize well to different game environment dimensions not seen during training.

\bibliographystyle{IEEEtran}
{\small
	\bibliography{references}}

\end{document}